# A Dataset and Baseline for Deep Learning-Based Visual Quality Inspection in Remanufacturing

Johannes C. Bauer*†, Paul Geng*, Stephan Trattnig*, Petr Dokládal†, and Rüdiger Daub*‡

* *Institute for Machine Tools and Industrial Management (iwb)*, *Technical University of Munich*, Garching, Germany
johannes.bauer@iwb.tum.de, paul.geng@iwb.tum.de, stephan.trattnig@iwb.tum.de
† *MINES Paris, PSL University*, *Centre for Mathematical Morphology (CMM)*, Fontainebleau, France
petr.dokladal@minesparis.psl.eu
‡ *Fraunhofer Institute for Casting, Composite and Processing Technology IGCV*, Augsburg, Germany
ruediger.daub@igcv.Fraunhofer.de

*Abstract*—Remanufacturing describes a process where worn products are restored to like-new condition and it offers vast ecological and economic potentials. A key step is the quality inspection of disassembled components, which is mostly done manually due to the high variety of parts and defect patterns. Deep neural networks show great potential to automate such visual inspection tasks but struggle to generalize to new product variants, components, or defect patterns. To tackle this challenge, we propose a novel image dataset depicting typical gearbox components in good and defective condition from two automotive transmissions. Depending on the train-test split of the data, different distribution shifts are generated to benchmark the generalization ability of a classification model. We evaluate different models using the dataset and propose a contrastive regularization loss to enhance model robustness. The results obtained demonstrate the ability of the loss to improve generalisation to unseen types of components.

*Index Terms*—remanufacturing, quality inspection, deep learning, gearbox, image data.

## I. INTRODUCTION

Remanufacturing describes the industrial process of restoring a used and worn product to like-new or sometimes even better-than-new condition [1]. Since a large part of the material and value-added in a product is retained [1], it promises ecologic as well as economic benefits [2]. After an initial quality assessment is performed, the returned products are first disassembled [3]. Then their components are cleaned and inspected to determine if they can be reused, or must be reworked or replaced. Finally, the product is reassembled and tested [4]. The included inspection procedures therefore have high impact on the quality of the remanufactured product as well as the economic feasibility of the process. Due to their complexity, these inspection tasks are usually carried out manually and rely heavily on expert knowledge. [3].

Deep neural networks (DNN) show great potential to automate such visual inspection tasks using image-based classification [5] or anomaly detection techniques [6]. However, most of the existing literature focuses on specific application scenarios and a small range of product variants or workpieces.

Due to the many uncertainties in remanufacturing, e.g., regarding the time and quantity of returned products and their condition [7], the developed automation systems should be able to handle a wide range of components and possibly unseen defect patterns. This poses a challenge for DNN that rely on comprehensive training datasets and usually assume identically and independently distributed (IID) inference data. If the visible failure patterns differ from those seen during training, the model's performance may significantly decrease. Such phenomena are also referred to as *dataset/distribution shift* [8]. Ideally, deployed models should be as robust as possible to such distribution shifts and still make correct or reliable predictions even if the data is not perfectly IID.

Currently, the development of methods to tackle these challenges is still hindered by a lack of publicly available, remanufacturing-specific image datasets that incorporate such forms of distribution shift. While different benchmark datasets for visual quality assurance exist [9], most of these focus on crack detection, e.g., [10], [11], or steel surface inspection, e.g., [12], [13]. Popular datasets to assess a model's robustness to distribution shifts stem from domains outside of manufacturing, like ImageNet-C [14] or WILDS [15], or tackle very specific problems, like printed circuit board inspection [16].

To contribute to the development of more robust and flexible DNN-based quality inspection methods, this work pursues two objectives. First, we propose a novel dataset for part inspection in remanufacturing. This dataset comprises images of gearbox components, like gear wheels or synchronizer parts (cf. Fig. 1), in working as well as defective condition, stemming from two different automotive transmissions. By splitting the data based on different criteria, distribution shifts of varying severity are generated to be able to benchmark a model's generalization capabilities more comprehensively. Second, a thorough evaluation of baseline classification approaches is performed with the proposed dataset. Our results show, that the generalization capability to unseen components and their defect patterns can be enhanced by contrastive regularization of the extracted embedding vectors.

The remainder of this paper is structured as follows: In section II, relevant fundamentals are explained and related works

Authors acknowledge funding from the German Federal Ministry of Research, Technology and Space under grant number 16KIS1805.

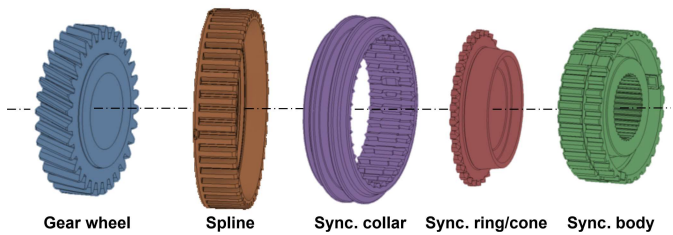

Fig. 1. Example renderings of parts from different component categories included in the dataset. CAD models are based on [17], [18] (not to scale).

on DNN-based visual quality inspection in remanufacturing are discussed. In section III, the proposed dataset is presented and characterized. This is followed by an evaluation of different DNN architectures and an assessment of the proposed regularization approach in section IV. Section V, concludes and gives an outlook on further research opportunities.

## II. FUNDAMENTALS AND RELATED WORKS

First, basic background knowledge about automotive transmissions and common wear patterns of relevant parts are outlined. Since the contained components (e.g., gear wheels) are material-intensive and costly to manufacture, they are well suited for remanufacturing and serve as an exemplary use case. Afterward, the state of the art regarding automated visual quality inspection in remanufacturing processes is discussed.

### A. Basics and Wear Patterns of Transmission Components

Automotive transmissions are complex products, containing many different components. They can be primarily classified into manual transmissions and automatic transmissions, although there exist different mixed and special forms [19].

*Structure:* In addition to the housing, shafts, bearings, seals, etc., the gears and the synchronizers are essential components of manual transmissions [19]. The gear pairs are arranged loosely on multiple shafts, with each shaft encompassing several gears. Depending on the desired transmission ratio, a different pair is connected to the shaft [19]. The synchronizer minimizes the speed difference between the shaft and the gear to be engaged before a gear change [20]. It includes the synchronizer cone, ring, body, and sleeve/collar [20]. Automatic transmissions typically use one or more planetary stages, with power transmission within these stages defined by various clutches and brakes [19]. The clutch packs and brakes occupy a significant portion of the transmission's volume. Interested readers may refer to [19] for more detailed explanations.

*Wear patterns:* The dataset focuses on gear wheels and other geared parts, like splines of clutch pack drums or synchronizer parts. Therefore, the presented wear patterns and defect types are restricted to these components. Typical gear wheel faults are summarized in the standard DIN 3979 [21]. These include surface fatigues like pitting (a breakout of small pieces of material), scratching, scoring, scuffing, cracks, different types of corrosion, and tooth breakage due to overload or fatigue (cf. samples 1 and 3 in Fig. 3). Proper functionality of the synchronizer requires a sufficiently high friction coefficient between the cone and the ring in order to align the speed of the idler gear and the shaft before their engagement [20]. If this is not given the speed equalization may fail, resulting in damage to other synchronizer components like the collar [20] (cf. samples 5, 6, and 8 in Fig. 3). The splines in the clutch packs may suffer from the axial movement of the clutch plates and discs, as well as the torsional stress when locking.

### B. Automated Visual Quality Inspection in Remanufacturing

Existing works on automated quality inspection in remanufacturing applications can be categorized according to their position in the remanufacturing process chain. They either deal with inspection of disassembled components or focus on the inspection of the whole product. Furthermore, the training process of the DNN can be a differentiating factor. A straightforward approach is the supervised training of classification or segmentation models. Alternatively, anomaly detection methods can be employed. Representative works are presented and discussed below.

*Supervised learning:* Nwankpa et al. were among the first to highlight the potential of DNN for quality inspection in remanufacturing [5]. They used the popular ResNet-18 architecture to classify eight different types of defects (e.g., pitting, rust, or cracks) on steel plates. A similar classification approach was used in [22] to detect samples with water stains after the cleaning process in remanufacturing of torque converters. Other works deal with localization or segmentation methods. This includes Saiz et al., who presented an ensemble model, combining the architectures DeepLabV3+ and YOLOv5, to detect defects on cages of constant velocity joints [23]. Similarly, Zheng et al. used a DNN-based localization method to inspect pipes. They employed the popular Mask-RCNN architecture to localize damages like holes or cracks in an image and map the pixel coordinates to an acquired 3D point cloud to aid automated planning of refurbishment processes [24]. More recently, Mohandas et al. presented an evaluation of different variants of YOLO and Mask-RCNN to detect and grade defects of smartphone displays [25].

Although the mentioned works demonstrate the potential of DNN for such inspection tasks, trained models are typically evaluated on randomly selected, identically distributed test data. This does not take the influence of distribution shifts into account. However, these are likely to occur in practice.

*Anomaly detection:* Kaiser et al. describe an anomaly detection method for inspection of automotive starter motors [26]. They train an autoencoder to reconstruct synthetically generated images of defect-free motors. Afterward, defects are detected by measuring the autoencoder's reconstruction error, which increases in the presence of defects. So far, the method has shown promising results on synthetically generated images but its application to realistic settings is still pending.

In practice, a higher degree of variation in the data might lead to additional challenges for the anomaly detection approach. Furthermore, inspecting multiple types of parts on a single station might pose difficulties. After all, detecting anomalies for multiple component types with a single model is much more difficult than for a single type of part [27]. In turn, using part-specific models to detect anomalies would require additional methods to classify their type first.

*Need for research:* To contribute to the practical application of automated visual quality inspection approaches in remanufacturing, developed inspection models should be able to handle a wide range of different components and failure types. This requires a high degree of generalization capability and robustness to distribution shifts, which should be evaluated on real-world data that explicitly includes examples of such distribution shifts. To date, no remanufacturing-specific benchmark datasets have been established that could be used for this purpose. Therefore, in this work, a novel dataset depicting gearbox components is proposed, which will be presented in the next section. Afterward, we evaluate different model architectures and propose a simple approach based on a contrastive regularization term to improve a model's generalization capability.

## III. PROPOSED DATASET

The dataset consists of images depicting components from two defective automotive transmissions, an automatic transmission previously installed in a *BMW E36* and a 6-speed manual transmission stemming from a *Mini r53*. Both gear boxes were first disassembled. The parts were then cleaned and assigned with part-specific IDs. In the following, the subsequent image acquisition and annotation are described in more detail, before the obtained dataset is characterized.

### A. Data Acquisition

Overall, 46 physical components were selected for image acquisition, with 19 stemming from the BMW and 27 from the Mini transmission. Since all selected components are nearly rotationally symmetrical, these were mounted on different 3D-printed axes with the camera facing the tooth flanks of the gear wheels or tooth-like features of the synchronizer components or splines, respectively (see also Fig. 1). The components were then incrementally rotated around the mounting axis to capture an RGB image of each tooth located in the focus region of the camera. The setup is schematically visualized in Fig. 2. This process was then repeated for the other side of the gear tooth, if it was not already visible from the first viewing angle. This means that a gear with 49 teeth results in 49 images of side A and another 49 images of side B. The camera model *U3-30C0CP-C-HQ Rev.2.2* of the company *IDS Imaging* equipped with a 35 mm lens was used for image acquisition. The setup featured two LED light panels for diffuse illumination. Since the components differ significantly in size and shape, patches of 128 to 128 pixels were extracted from the focus region, as visualized in Fig. 2. Thereby, the number of total images was further increased and overlapping regions within images of a single component could be avoided.

### B. Data Annotation

This section first discusses the annotation criteria before the iterative labeling process is explained. Finally, the split of the data into different sets of training and test data is discussed.

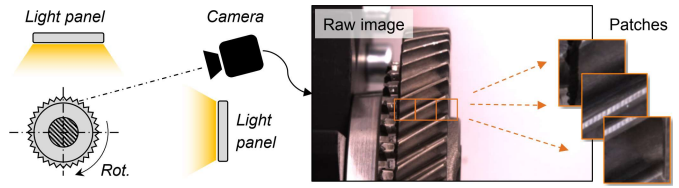

Fig. 2. Schematic visualization of the acquisition setup and subsequent patch extraction step.

*Annotation criteria:* The acquired image patches were afterwards manually assigned to one of the two classes *OK* and *nOK* (not OK) according to the visible signs of wear. Thereby, small and sporadic signs of pitting or light scratches were classified as still OK, while images showing larger pitting, scratches, scoring, or scuffing were classified as nOK. Deformations of the component, like notches on the contours, or clearly ground-in or corroded surfaces, were classified as nOK. Stains from the cleaning process or small oil residues were instead classified as OK. Edge cases were discussed between the authors. Image patches showing no relevant regions were discarded. Overall, 15,588 annotated image patches were obtained. Exemplary samples are visualized in Fig. 3.

*Annotation improvement:* To improve the label quality and remove the risk of label errors, so-called *confident learning* [28] approaches were employed in two steps. In a first step, the images were randomly split into five cross-validation (CV) folds and an ensemble model was trained on the training data of each fold and afterward applied to its test set. The false model predictions were gathered and their labels were manually checked to identify samples that were given a false label by accident. The procedure was repeated until no more annotations were identified that should be changed. During the process, the number of false model predictions decreased from 1,161 to 543, 422, and 390. The remaining false predictions are mainly attributed to class imbalances within images of specific components, which is further discussed in sections III-C and IV-A. The second step aimed to reevaluate labels near the decision boundary where samples are hard to classify for a human annotator as well as for a DNN. Therefore, model predictions with a confidence score (corresponding to the maximum softmax probability) below 0.75 were selected, regardless of whether they were classified correctly or not. These 635 samples were manually reviewed, and 106 labels were changed.

*Construction of splits:* Distribution shifts caused by new products, components, or unknown defect patterns are a major challenge for quality inspection in remanufacturing. To be able to evaluate models and methods with regard to such, we construct distribution shifts of varying severity. To do so, relevant metadata (e.g., part ID or component type/category) were gathered and used to create different train-test splits. For each split, annotations to perform 4- or 5-fold CV were exported.

S1 A simple option is to *randomly* split all images. This results in very low discrepancy between training and test set, since both contain images from all parts and acquisition settings etc.

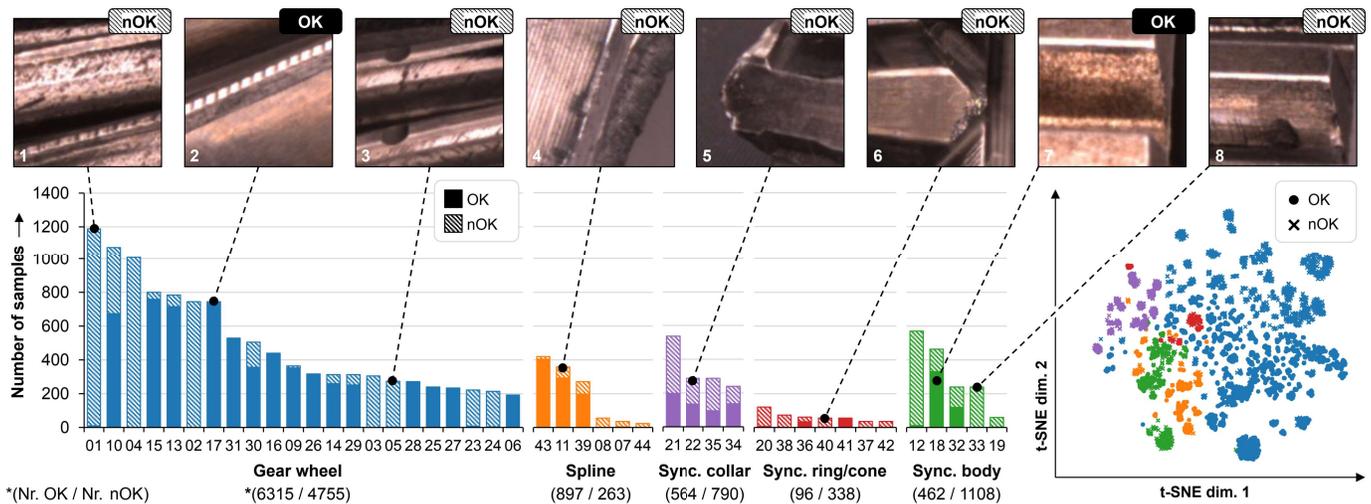

Fig. 3. Overview of the dataset, incl. exemplary images and the label distribution for individual functional part IDs (01, 02, . . . , 44) and component categories (gear wheel, spline, synchronizer collar, etc.). Colors correspond to the t-SNE representations of the embedded images shown in the lower right (best viewed in color).

- S2 A second option is to split using the ***acquisition settings***. Thereby, the training and test sets might include images from the same component but not from the same side of its gear teeth. Failure patterns are therefore less similar.
- S3 A third option is to split among physical parts. However, some parts are still identical, such as the gear wheels of the planetary stages. Therefore, so-called ***functional part IDs*** were assigned, where identical parts have the same ID. Functionally integrated parts (e.g., gear wheels with attached synchronizer cones as seen in Fig. 2) received multiple functional part IDs, effectively treating them as if they were separated parts.
- S4 Lastly, we split based on the ***component category***. Therefore, each functional part ID was assigned to one of the 5 categories *gear wheel*, *synchronizer ring/cone*, *synchronizer body*, *synchronizer collar*, and *spline* (cf. Fig. 1). In 4 iterations, the synchronizer rings/cones, bodies, collars, and the splines are each used for testing once, while the remaining 4 categories are used as training data. The gear wheels are not used for testing but are included in the training data of every CV fold as they make up the majority of samples in the dataset.

The obtained dataset will be characterized in more detail in the next section.

### C. Data Characterization

The final label distribution is visualized in Fig. 3. It depicts the number of OK and nOK images per functional part ID (S3) and component category (S4) as well as exemplary images. Two-dimensional sample representations – generated using the embedding vectors of an ImageNet-pretrained ResNet-50 and the dimensionality reduction technique t-SNE (t-distributed stochastic neighbor embedding) [29] – indicate the visual difference between the samples.

Overall, the dataset is well balanced, including 8,334 OK and 7,254 nOK images. Most images, about 71 %, stem from gear wheels. When focusing on single functional part IDs, the data is less balanced and the majority of parts are either mostly OK or mostly nOK. Nevertheless, each component category includes samples from both classes. The highest imbalance is observed for the synchronizer rings/cones, where only about 22 % of samples are labeled as OK.

### IV. BASELINE EVALUATION

Since all images are assigned to one of the two classes OK and nOK, a supervised classification is performed to establish a baseline. In the following, we first report on the evaluation of different model architectures applied to the proposed splits. Afterward, the improved baseline using contrastive regularization on the extracted feature embeddings is presented and evaluated.

### A. Evaluation of Model Architectures and Distribution Shifts

As a first step, different models were applied to the four different splits to assess their generalization performance and quantify the distribution shifts' severity. After reporting on the methodology and implementation, the obtained results are presented and discussed.

*Method:* We consider both Convolutional Neural Networks (CNN) and Transformer-based model architectures. The models are either trained completely or finetuned, freezing the feature extractors' weights. The evaluated model architectures include the CNN architectures ResNet-50 [30], DenseNet-121 [31], MobileNetV3-S [32], and EfficientNetV2-M [33] as well as the Transformer-based architectures SwinV2-B [34], and DINOv2 with registers [35]–[37]. DINO further differentiates itself by its self-supervised pretraining process based on large amounts of image data. The remaining models are all initialized with weights pretrained on ImageNet1k. The implementations are based on PyTorch's torchvision package [38] and the Huggingface transformer library [39].

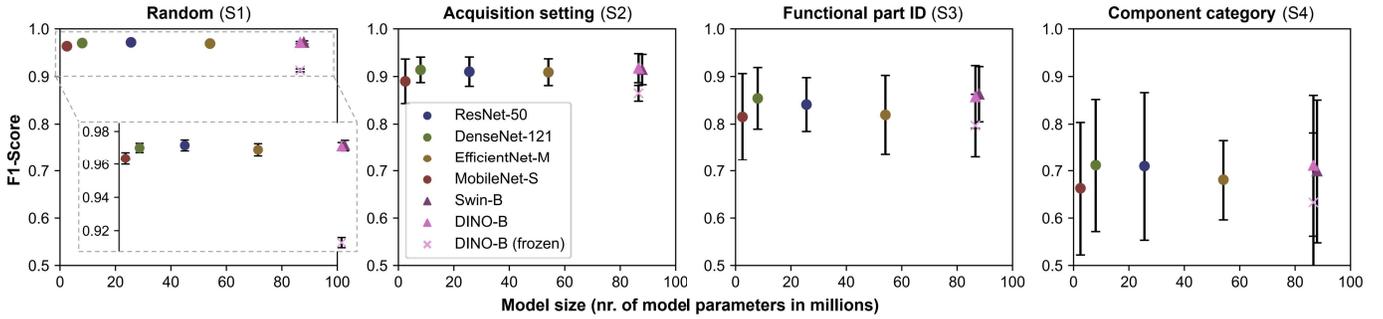

Fig. 4. Average F1-Scores of different models obtained during the classification task using different split configurations (S1 to S4). The error bars correspond to +/- one standard deviation across the CV folds (best viewed in color).

We employ a batch size of 128 and utilize random horizontal and vertical flips as well as Gaussian noise for data augmentation. Images are normalized by mean $\mu = (0.485, 0.456, 0.406)$ and standard deviation $\sigma = (0.229, 0.224, 0.225)$. The Adam optimizer is used in combination with a cosine annealing learning rate scheduler. The learning rate and the number of epochs are tuned for each architecture. Due to the class imbalance of individual splits, the F1-Score is reported as a performance metric. It represents the arithmetic mean of precision and recall. For each data split (S1 to S4), a 5-fold (4-fold in case of S4) CV is performed.

***Results:*** The classification results are visualized in Fig. 4, comparing the obtained mean F1-Score, its standard deviation, and the model size for the four split configurations S1 to S4. No major performance differences between the models are observed. On the random split (S1), all fully trained models achieve an average F1-Score of over 96 % with a standard deviation below 0.4 %. For the other splits, performance decreases and the standard deviation increases. When the data is split based on the component category (S4), the mean F1-Scores range between 63 % and 71 %. The DINO models with frozen feature extractors perform slightly worse than the fully trained versions. Its average F1-Scores range between 91.3 % on split S1 and 63.3 % on split S4.

***Discussion:*** Although the distribution shift between training and test sets is minimal in case of the random split (S1), the models do not achieve an F1-Score above 97.2 %. As mentioned earlier, this may be attributed to class imbalances within images of specific components (cf. Fig. 3). For some components there are only few images of defects. Hence, these defect patterns are rare and difficult to classify for the model. Furthermore, if components have a predominant class, models might tend to base their decision on component features and not on defect patterns. This would lead to false predictions for the deviating image samples of such a component. Overall, the decline in performance and the increase in standard deviation suggest an increasing distribution shift with increasing split numbers (S2 to S4).

Regarding the model architectures, the use of larger Transformer models alone does not appear to provide any benefits. This is somewhat unexpected since these larger models are usually associated with improved robustness and generalization capability. On the one hand, the individual differences of samples may be too large or too small for these effects to play a significant role. On the other hand, an influence of suboptimal hyperparameter values cannot be completely ruled out, despite the efforts for hyperparameter tuning invested in this work. The worse performance of the DINO models with frozen feature extractors indicates a domain gap between DINO's pretraining data and this dataset.

### B. Improving Generalization With Contrastive Regularization

The integration of the proposed contrastive regularization was motivated by the observation that the differences between defect patterns (e.g., scratches or scores) are usually smaller for different categories of components than the overall appearance of those components. Furthermore, since all of the images of a component are often either predominantly OK or predominantly nOK, models may be prone to learning false correlations. The approach presented and evaluated in the following aims to make the model learn more component-independent features.

***Method:*** The introduced regularization builds on methods from the field of *contrastive representation learning*. These methods learn an embedding function, the DNN, by maximizing the similarity of an extracted embedding vector to those of the same category and minimizing the similarity with representations of different categories [40]. Some operate in a self- or semi-supervised manner, treating only a single sample and an altered version of it as a positive pair and all other samples in a mini-batch or queue as negatives, e.g., [41], [42].

In our scenario, every sample from the same class (OK/nOK) could be treated as a positive sample, regardless of the depicted component and its category, and samples from the other class would be treated as negatives. This way, representations of OK samples should move closer together in the embedding space while moving further away from the nOK samples and vice versa.

To integrate this objective into the training process, an additional loss term, based on the *soft nearest neighbor loss*

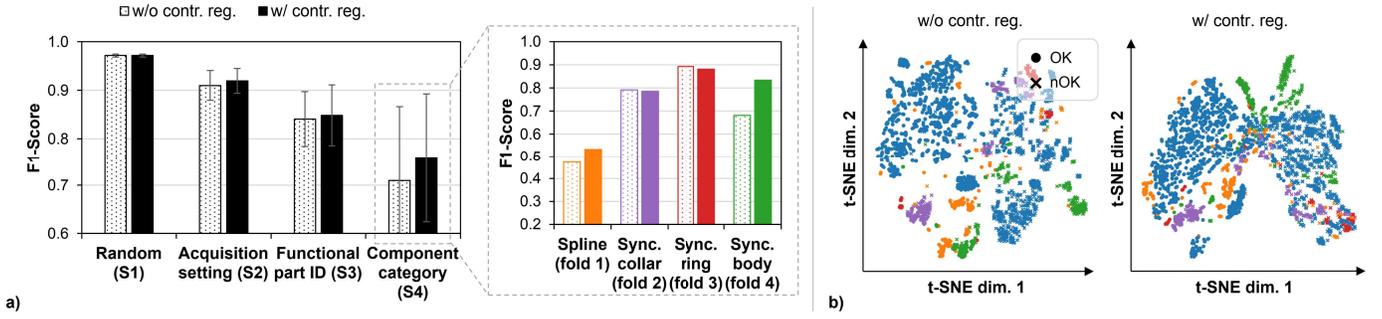

Fig. 5. Average F1-Score of a ResNet-50 trained with (w/) and without (w/o) contrastive regularization (a). Error bars indicate +/- one standard deviation across the CV folds. Results for the individual CV folds of split S4 are further detailed to the right. The bar colors correspond to the t-SNE visualization of the embedding vectors shown in (b). These were computed on the test data of the first fold of the randomly split data (S1) (best viewed in color).

proposed by [43] and [44], is applied. The loss for a batch of $N$ processed samples $(y, \hat{y}, v)$ is calculated as stated in Eq. 1,

$$loss = l_{ce}(y, \hat{y}) + \alpha \cdot l_{sn}(v, y, T) \quad (1)$$

where $\alpha$ and $T$ are scalar hyperparameters, $l_{ce}(y, \hat{y})$ represents the cross-entropy loss function, $y$ are the true labels, $\hat{y}$ are the model predictions, and $v$ are the extracted embedding vectors before the model's final classification layer. The soft nearest neighbor loss $l_{sn}$ is calculated as stated in Eq. 2,

$$l_{sn}(v, y, T) = -\frac{1}{N} \sum_{i \in 1..N} \log \left( \frac{\sum_{\substack{j \in 1..N, \\ i \neq j, \\ y_i = y_j}} e^{-d(v_i, v_j)/T}}{\sum_{\substack{k \in 1..N, \\ k \neq i}} e^{-d(v_i, v_k)/T}} \right) \quad (2)$$

where $d()$ is the cosine distance.

After an initial exploration of suitable hyperparameters on the first fold of S3, $\alpha$ was set to 0.2 and $T$ to 2.0. Afterward, a ResNet-50, trained using the additional loss term, was evaluated on all four split configurations. We selected this model architecture since it is widely used and constitutes a popular baseline model for scientific studies. The learning rate and number of epochs were tuned following the same procedure as for the other models. The remaining hyperparameters remained unchanged to section IV-A.

*Results:* The obtained classification results are visualized in Fig. 5 a). It is noticeable that the model trained using the additional regularization term shows an improved mean F1-Score for all four split configurations. The improvement rises with increasing distribution shift. While it is only marginal on the random split (S1), it accounts for approximately one percentage point for splits S2 and S3. When splitting using the component category (S4), the mean F1-Score is increased from approximately 71 to 76 %. A closer look at the individual components of the CV folds in split S4 shows that the improvement can be mainly attributed to the synchronizer body (green). This contrasts with the spline components (orange), to which both models do not generalize well. For the synchronizer collars and rings, performance has deteriorated slightly by approximately one percentage point.

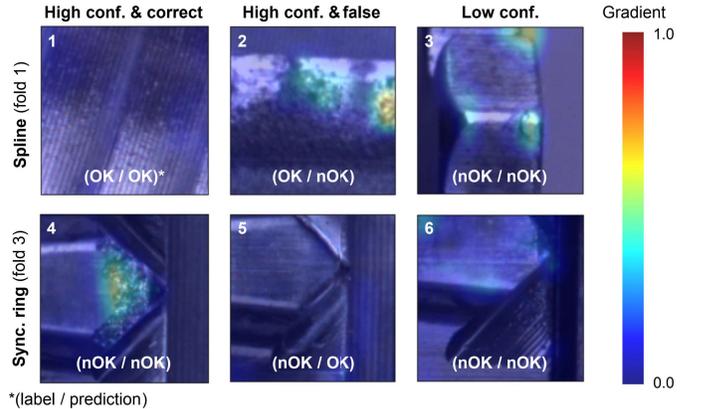

Fig. 6. GradCAM visualizations for exemplary samples in the CV folds 1 and 3 of split S4.

Fig. 5 b) displays t-SNE representations of the extracted embeddings for models trained with and without contrastive regularization. The stated marker labels are based on the ground truth annotations. In both cases two mainly distinct regions for nOK and OK samples are formed. However, the contrastive loss is reflected in the more pronounced margins of the plot on the right.

*Discussion:* While the contrastive loss can improve the average F1-Scores, the performance still drops significantly as soon as training and test data become less similar. Especially on the splits S3 and S4, there is still potential for further improvement. Moreover, the generalization performance strongly varies across specific component IDs and categories. This is already indicated by the F1-Scores' high standard deviation over the CV folds and is also evident in the test results for individual component categories in Fig. 5 a).

To investigate potential reasons for the observed generalization gaps, GradCAM (Gradient-weighted Class Activation Mapping) [45] visualizations of selected samples are depicted in Fig. 6. GradCAM is a popular method for visually explaining DNN by inspecting the gradients in hidden layers. We used the last convolutional layer of the third residual block of the ResNet-50 trained with the proposed contrastive loss on folds 1 and 3 of split S4. The implementation is based on [46]. Gradient values were scaled using the same value for all samples. To inspect different failure cases of the model, we explicitly selected samples where the model predicted with high confidence and was either wrong/correct, as well as samples where the model predicted with low confidence.

Two failure modes are evident that may contribute to a reduced generalization performance. On the one hand, features of wear or defect patterns may be present in the image that the model does not recognize. This should be indicated by equally low gradients over the whole image. In case the sample is indeed OK, such as sample 1, this is no problem. However, it may be the cause for the false assessment of sample 5. On the other hand, the model may recognize features that are associated with defects on the training data but not on the test data. This specific form of distribution shift is also known as concept drift [47]. While learned defect features may have been correctly generalized to samples 3 and 4, the surface irregularities of the slightly worn cast component in image 2 were found to be still OK in this case.

## V. Conclusion

*Summary:* In this work, a new dataset for the evaluation of visual quality inspection approaches in the remanufacturing domain was presented. By applying different split configurations (S1 to S4) to the acquired image data, distribution shifts of varying severity between the respective training and test datasets were generated. Subsequently, various model architectures were evaluated and a custom regularization approach was presented that was shown to improve generalization robustness. Since the obtained performance strongly varies depending on the split configuration, the dataset offers the opportunity to evaluate novel models and methods more thoroughly than conventional IID scenarios. We hope this contributes to the development of more flexible and robust approaches that ultimately meet the practical challenges of automated quality inspection in remanufacturing.

*Limitations:* The dataset comprises over 15,000 images of real-world components showing actual wear and defects. However, the images are derived from 46 physical parts in two gearboxes. While this is not enough to capture the full spectrum of potential defect types or components, the different splits at least allow to test how a model or method performs when facing unknown parts. Depending on the selection of the test components, the obtained performance can strongly vary. Therefore, performing CV is crucial, although it is computationally expensive. Furthermore, the decision on whether a sample should be rated as OK or nOK is based on subjective criteria. Different individuals might come to different conclusions, especially since many samples represent borderline cases, posing an additional challenge for both humans and DNN. Nevertheless, precisely these challenges should facilitate the development of novel approaches and algorithms.

*Future research opportunities:* We focused on improving generalization capability and investigated a regularization-based approach. Data-centric approaches could also improve performance on the presented dataset, e.g., by incorporating data from similar datasets into the training data or generating synthetic images. Besides the classification performance, model calibration and the extent to which generated confidence values or uncertainty measures indicate whether a prediction is likely to be false also play an important role [48]. Their robustness could be assessed on this data as well. Additionally, approaches for quick and efficient adaptation of DNN to new components and/or novel defect patterns, e.g., [49]–[52], could be evaluated and further extended utilizing the proposed dataset, too. Lastly, the image data could be utilized to also evaluate anomaly detection or segmentation methods.

## Data Availability

The presented dataset can be accessed via: https://www.kaggle.com/datasets/jhnnsbr/gearbox-components-remanufacturing-inspection.

## Author Contributions

JCB: Conceptualization, Methodology, Software, Investigation, Data Curation, Writing - Original Draft, Writing - Review & Editing, Visualization, Project administration. PG: Conceptualization, Methodology, Data Curation, Writing - Review & Editing ST: Methodology, Writing - Review & Editing. PD: Methodology, Supervision, Writing - Review & Editing. RD: Resources, Supervision, Funding Acquisition, Writing - Review & Editing.